%% file: sn-article.tex
\documentclass[11pt, a4paper, logo, nonumbering]{style}
\usepackage[numbers, sort&compress]{natbib}
\usepackage{dblfloatfix}
\usepackage{caption}
\usepackage{xspace}
\usepackage{pifont}
\usepackage{multirow}
\usepackage{tcolorbox}
\usepackage{longtable}
\usepackage{hyperref}
\usepackage{amsfonts}
\usepackage{amsmath}
\usepackage{amssymb}
\usepackage{adjustbox}

\usepackage[bottom]{footmisc}
\usepackage{listings}
\usepackage{setspace}

\usepackage{array}
\usepackage{tabularx}
\usepackage{booktabs}

\usepackage{multicol}

\newcommand{\answerTODO}[1][]{\textcolor{red}{[TODO]}}

\definecolor{codebackground}{rgb}{0.95,0.95,0.95}
\definecolor{codeframe}{rgb}{0.8,0.8,0.8}
\definecolor{keyword}{rgb}{0.0,0.0,0.55}
\definecolor{stringcolor}{rgb}{0.65,0.13,0.13}
\definecolor{commentcolor}{rgb}{0.0,0.5,0.0}

\lstdefinestyle{code}{
    backgroundcolor=\color{codebackground},
    frame=single,
    rulecolor=\color{codeframe},
    basicstyle=\ttfamily\footnotesize,
    keywordstyle=\bfseries,
    stringstyle=\color{stringcolor},
    commentstyle=\itshape\color{commentcolor},
    showspaces=false,
    showstringspaces=false,
    showtabs=false,
    tabsize=4,
    captionpos=b,
    breaklines=true,
    breakatwhitespace=true,
    sensitive=true,
    breakindent=0pt,
}

\makeatletter
\def\@BTrule[#1]{%
  \ifx\longtable\undefined
    \let\@BTswitch\@BTnormal
  \else\ifx\hline\LT@hline
    \nobreak
    \let\@BTswitch\@BLTrule
  \else
     \let\@BTswitch\@BTnormal
  \fi\fi
  \global\@thisrulewidth=#1\relax
  \ifnum\@thisruleclass=\tw@\vskip\@aboverulesep\else
  \ifnum\@lastruleclass=\z@\vskip\@aboverulesep\else
  \ifnum\@lastruleclass=\@ne\vskip\doublerulesep\fi\fi\fi
  \@BTswitch}
\makeatother

\addto\extrasenglish{
}

\renewcommand{\today}{}

\title{\centering  Claw-SWE-Bench: A Benchmark for Evaluating OpenClaw-style Agent Harnesses on Coding Tasks}

\renewcommand\Affilfont{\normalfont\small\centering}
\author[1]{\mbox{Mengyu Zheng}}
\author[1]{\mbox{Kai Han}}
\author[2]{Boxun Li}
\author[2]{Haiyang Xu}
\author[5,1]{\mbox{Yuchuan Tian}}
\author[1]{Wei He}
\author[1]{Hang Zhou}
\author[3]{Jianyuan Guo}
\author[1]{Hailin Hu}
\author[4]{Lin Ma}
\author[5]{Chao Xu}
\author[6,2]{Guohao Dai}
\author[2]{Lixue Xia}
\author[7]{\mbox{Yunchao Wei}}
\author[1]{\mbox{Yunhe Wang}}
\author[8]{Yu Wang}
\affil[1]{\mbox{TokenRhythm Technologies}}
\affil[2]{\mbox{Infinigence AI}}
\affil[3]{\mbox{City University of Hong Kong}}
\affil[4]{\mbox{SEE Fund}}
\affil[5]{\mbox{Peking University}}
\affil[6]{\mbox{Shanghai Jiaotong University}}
\affil[7]{\mbox{Beijing Jiaotong University}}
\affil[8]{\mbox{Tsinghua University}\protect\\[3pt]\protect\normalfont\footnotesize\texttt{\{mengyu.zheng,\,kai.han,\,yunhe.wang\}@tokenrhythm.ai}\quad\texttt{yu-wang@mail.tsinghua.edu.cn}}

\input{commands.tex}

\input{content_compatibility.tex}

\input{sections/00_abstract}

\begin{document}
\maketitle
\input{sections/01_introduction}
\input{sections/06_related_work}
\input{sections/02_clawswebench}
\input{sections/03_lite}
\input{sections/04_experimental_setup}
\input{sections/05_results}
\input{sections/07_conclusion}
\label{page:main_end}

\clearpage
\bibliography{sn-bibliography}

\clearpage
\input{appendix_content}

\end{document}

%% file: commands.tex
\renewcommand{\phi}{\varphi}

\renewcommand{\epsilon}{\varepsilon}
\renewcommand{\imath}{\mathrm{i}}

\newlength{\restsubwidth}
\newlength{\restsubheight}
\newlength{\restsubmoreheight}
\newcommand{\rest}[2]{%
        \settowidth{\restsubwidth}{\ensuremath{#2}}
        \settoheight{\restsubheight}{\ensuremath{{}_{#2}}}
        \ensuremath{{#1\hskip 0.5pt}_{\vrule\kern2pt\parbox[b][%
        4pt][b]{\the\restsubwidth}{%
                        \ensuremath{{}_{#2}}}}}
        }

%% file: content_compatibility.tex
\usepackage{wrapfig,capt-of,fvextra}
\RecustomVerbatimEnvironment{verbatim}{Verbatim}{
  fontsize=\small,
  breaklines=true,
  breakanywhere=true,
  breaksymbolleft={},
  breaksymbolright={}
}

%% file: sections/00_abstract.tex
\begin{abstract}
The software engineering capabilities of general-purpose agent harnesses remain underexplored, and existing benchmarks offer limited support for comparing these harnesses under consistent conditions.
To address this gap, we introduce Claw-SWE-Bench, a unified benchmark that enables researchers to systematically assess the capabilities and efficiency of general-purpose harnesses on software engineering tasks.
The benchmark contains 350 real-world GitHub issue-resolution instances across eight programming languages and 43 repositories and provides a shared adapter protocol to align task inputs, outputs, and execution environments across harnesses.
Experiments show that general-purpose harnesses can effectively resolve real-world software issues and that their success rates and resource consumption vary substantially even when the underlying model is held fixed.
To lower evaluation costs and support faster debugging and iteration, we also provide Claw-SWE-Bench Lite, an 80-instance subset designed to preserve the key evaluation properties of the full benchmark.
We hope this benchmark will help researchers better evaluate and understand the performance of general-purpose harnesses on software engineering tasks and guide the development of more capable and efficient harnesses.
\input{project_links.tex}
\end{abstract}

%% file: project_links.tex
The data is available at \url{https://github.com/opensquilla/claw-swe-bench} and \url{https://huggingface.co/datasets/TokenRhythm/Claw-SWE-Bench}.

%% file: sections/01_introduction.tex
\section{Introduction}

Software development and maintenance are important application areas for agents.
Automating tasks such as code generation and issue resolution could reduce the workload for developers and improve development and maintenance efficiency~\citep{hou_llm4se_survey_2024}.
However, real-world software engineering tasks require more than code generation.
Agents must understand problem descriptions, locate relevant code, and verify the correctness of changes through execution and testing~\citep{jimenez_swebench_2024,swe_agent}.
To address these challenges, prior work has used methods such as supervised fine-tuning~\citep{swe_smith} and reinforcement learning~\citep{luo_deepswe_2025} to improve the software engineering capabilities of the underlying models.

Task completion depends not only on the capabilities of the underlying model but also on the harness~\citep{swe_agent,lee_meta_harness_2026}.
This execution framework coordinates interactions between the model and the external environment.
In recent years, agent harnesses, particularly general-purpose harnesses such as OpenClaw~\citep{steinberger_openclaw}, have continued to evolve and gain widespread adoption.
They organize information about the environment, provide tool interfaces, and manage interaction context and execution workflows.
These functions support models in tasks involving productivity tools, browser automation, computer use, and scientific assistance~\citep{ding_wildclawbench_2026,generic_agent_2026}.
Existing benchmarks and practical applications have demonstrated the strong performance of general-purpose harnesses on everyday tasks~\citep{li_clawsbench}.

Despite their strong performance in these settings, the ability of general-purpose harnesses to handle software engineering tasks has received little systematic study.
More importantly, it remains unclear whether general-purpose harnesses with different designs perform differently on coding tasks.

In this work, we introduce Claw-SWE-Bench, a benchmark for studying the performance of general-purpose harnesses on coding tasks.
It contains 350 real-world GitHub issue-resolution instances from the SWE-bench family~\citep{swe_smith,swebench_verified_mini}, covering eight programming languages and 43 repositories.
These tasks allow us to compare and evaluate how harnesses affect coding performance.
Many existing evaluations of agent coding capabilities focus primarily on models~\citep{deng_swebench_pro,huang_deepswe_2026}.
A platform for fair comparisons of different general-purpose harnesses under consistent conditions is still lacking.
Consistent with resource-aware agent evaluation~\citep{kapoor_hal,fan_swe_effi}, Claw-SWE-Bench reports task resolved rate , API cost, and average runtime to characterize both the task-solving capabilities and execution efficiency of general-purpose harnesses.

Claw-SWE-Bench provides a shared adapter protocol to align task inputs, outputs, and execution environments.
The protocol integrates different harnesses into a unified evaluation pipeline.
It standardizes task environments, outer runtime budgets, patch extraction, and test-based evaluation.
We identify a source of information leakage in some task environments: Git commits after the specified base commit remain accessible.
We remove these commits before evaluation to prevent agents from inspecting subsequent fixes.
Using Claw-SWE-Bench, we systematically evaluate seven general-purpose harnesses with three underlying models.
We find that general-purpose harnesses can perform software development tasks effectively.
We also find substantial differences in task success rates across general-purpose harnesses, even when the underlying model is held fixed.
These results show the importance of harness design for agent performance.
In terms of efficiency, token usage also varies substantially across harnesses.
This indicates that harness design affects both task performance and resource consumption during execution.

Finally, we construct Claw-SWE-Bench Lite, a compact evaluation subset containing 80 instances, to simplify evaluation and iteration for agent developers and reduce debugging costs.
Its construction accounts for language coverage and the task difficulty distribution.
Lite balances agreement with the full benchmark in task success rates, relative performance across harness--model pairs, and cost characteristics.
This preserves the key evaluation properties of the full benchmark at lower cost.

Through Claw-SWE-Bench, we establish a unified benchmark for evaluating the software engineering capabilities of harnesses.
We also show that general-purpose harnesses have strong coding capabilities.
We hope that the benchmark and its Lite subset will help researchers and developers evaluate and refine agents, supporting progress in agent research and applications.

%% file: sections/06_related_work.tex
\section{Related Work and Background}\label{sec:related}

Many existing benchmarks for general-purpose agents focus on assistant tasks~\citep{mialon_gaia_2024,yoran_assistantbench_2024,drouin_workarena_2024}.
Many evaluations of software engineering agents emphasize model comparisons, with less systematic attention to the effects of harness design~\citep{deng_swebench_pro,huang_deepswe_2026}.
Taken together, these settings provide limited evidence on performance differences among general-purpose harnesses on real-world software engineering tasks~\citep{kapoor_agents_matter_2025,yehudai_agent_evaluation_survey_2026}.
For example, ClawsBench~\citep{li_clawsbench} evaluates agent capabilities and safety in simulated productivity workflows, including email management and scheduling.
WildClawBench~\citep{ding_wildclawbench_2026} compares harnesses across a range of assistant tasks.
Its coding scenarios, such as writing inference scripts for SAM3 and reproducing benchmark results, are manually authored rather than drawn from real GitHub issues and their corresponding fixes.
DeepSWE~\citep{huang_deepswe_2026} primarily compares model capabilities under a fixed harness rather than systematically evaluating the role of harness design.


\paragraph*{General-Purpose Agent Harnesses.}

General-purpose harnesses~\citep{wu_autogen_2024,gao_agentscope_2024,wang_autonomous_agents_survey_2024} connect language models to external environments and support tool use across applications.
They construct model context from environmental observations and interaction history, execute tool calls, and maintain session state.
OpenClaw~\citep{steinberger_openclaw} adopts a gateway architecture that connects messaging channels to session-based agent execution.
Its runtime assembles workspace instructions and reusable skills into the prompt, coordinates tool calls, and persists interaction histories.
Hermes Agent~\citep{nous_hermes_agent} emphasizes reuse across sessions through persistent memory and a skill library.
It retains environment facts and user preferences, retrieves past conversations, and stores task procedures as skills that can be loaded and revised when needed.
nanobot~\citep{hkuds_nanobot} takes a lightweight approach, organizing a compact Python agent loop around model calls, tool execution, and message handling.
More generally, memory and reusable skills can be supplied as context, while file operations, shell commands, and web tools support tasks across applications~\citep{wang_workflow_memory_2025,wu_os_copilot_2024}.
ZeroClaw~\citep{zeroclaw_labs} emphasizes modularity through a Rust runtime with trait-based interfaces for model providers, tools, communication channels, and memory backends.
These interfaces allow components to be replaced independently, while the runtime manages the agent loop and enforces execution policies.



%% file: sections/02_clawswebench.tex
\section{Claw-SWE-Bench}
\label{sec:claw_swe_bench}

We introduce Claw-SWE-Bench to study the performance of general-purpose agent harnesses, referred to here as \emph{claws}, on real-world software engineering tasks. Following the SWE-bench task setting~\citep{jimenez_swebench_2024}, agents modify existing repositories to resolve real GitHub issues, and the resulting patches are evaluated through repository-level tests. The benchmark contains 350 task instances from SWE-bench-Multilingual~\citep{swe_smith} and SWE-bench-Verified-Mini~\citep{swebench_verified_mini}, spanning 43 repositories and eight language groups: Java, Go, Rust, JavaScript/TypeScript, C/C++, Ruby, PHP, and Python. This coverage across languages and repositories broadens the range of software engineering settings represented in the evaluation.

Evaluating these claws requires both reliable integration and consistent conditions for comparison. We first provide an adapter protocol that connects native harness execution to the repository editing and patch submission required by SWE-bench. This allows general-purpose harnesses to participate in the evaluation while preserving their internal agent loops. Building on this protocol, we define a standardized execution pipeline that controls task inputs, initial repository states, runtime conditions, patch extraction, and test-based scoring across harnesses. Different harnesses can then be compared with the underlying model and outer execution budget held fixed. Their internal tool interfaces and execution strategies remain part of the harness design under evaluation.

\begin{figure}[t]
  \centering
  \includegraphics[width=0.9\linewidth]{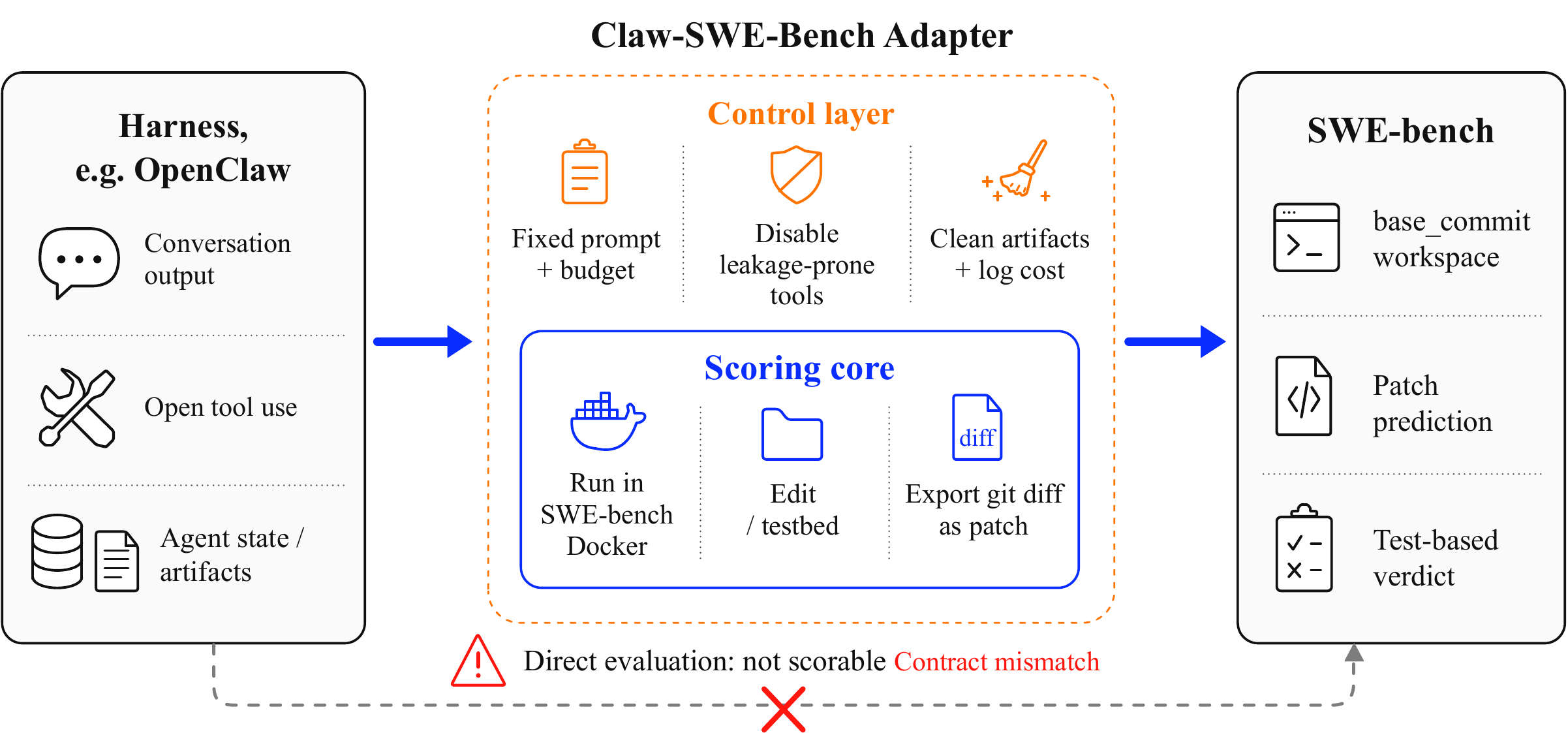}
  \caption{Contract mismatch between OpenClaw-style harnesses and SWE-bench.
The adapter converts a general agent interaction into a SWE-bench-scored patch prediction, while outer controls ensure fairness, comparability, and traceable cost.}
  \label{fig:framework_pipeline}
  \vspace{-3mm}
\end{figure}

\subsection{The Adapter Protocol}
\label{sec:adapter_protocol}

The core artifact of \textsc{Claw-SWE-Bench} is a harness-agnostic adapter protocol. Every supported harness implements the same abstract interface:
\texttt{create\_agent},
\texttt{send\_task},
\texttt{backup\_session},
\texttt{delete\_agent},
\texttt{get\_docker\_args}.
The shared orchestrator calls these methods without knowing which harness is underneath. This design separates the benchmark lifecycle from the agent implementation: container management, prompt rendering, patch collection, prediction writing, metadata logging, resume support, and evaluation are implemented once, while each harness adapter only supplies the code needed to invoke its own agent loop inside the container.

At run time, the shared orchestrator owns the common lifecycle shown in Figure~\ref{fig:framework_pipeline}: container startup, repository reset, prompt rendering, patch collection, prediction writing, metadata logging, and evaluation. The adapter supplies only the harness-specific hooks needed to create or configure an agent, dispatch the rendered task, preserve any run artifacts, and clean up harness state. This boundary is deliberate: the benchmark controls the task-facing environment and evaluator-facing patch format, while leaving the internal agent loop to the harness under study.

Crucially, the candidate patch is collected by the benchmark runner rather than parsed from the agent's final message. Agents communicate a solution only by editing files in the repository. This makes the output contract independent of whether a harness emits JSON, plain text, a final answer, or no structured response at all.

All harnesses are launched through a single command-line entry point, \texttt{run\_infer.py}. The user specifies the harness name, dataset configuration, model identifier, run identifier, timeout, worker count, and optional instance filters. Dataset metadata are loaded from the configured SWE-bench source, and each instance is represented by the fields needed by the protocol: \texttt{instance\_id}, \texttt{repo}, \texttt{base\_commit}, and \texttt{problem\_statement}. A harness registry maps a string identifier to the corresponding adapter class; adding a new claw therefore requires only implementing the adapter interface and registering it in the harness map---the dataset loader, Docker workspace manager, prompt builder, patch collector, prediction writer, and evaluator remain untouched.

\subsection{Standardized Execution Pipeline}
\label{sec:exec_pipeline}

Around the adapter boundary, \textsc{Claw-SWE-Bench} standardizes the parts of the evaluation stack that would otherwise confound harness comparisons.

\textbf{Runtime and workspace.}
Each task runs inside its corresponding SWE-bench evaluation Docker image~\citep{jimenez_swebench_2024}, with the repository reset to the instance's \texttt{base\_commit} under \texttt{/testbed}. To prevent agents from retrieving existing solutions online, we block external network access during task execution, except for connections required for model inference. The same outer budget is used for every harness: \texttt{max\_turns}=300 where supported, a 3600-second wall-clock timeout, and fixed worker concurrency. Harness-specific dependencies are supplied through Docker arguments or bind mounts, but the repository state, evaluation image, and outer budget seen by the agent are fixed.

\textbf{Prompt rendering.}
Every instance is rendered with the same task prompt. The prompt contains the problem statement, identifies \texttt{/testbed} as the code repository, and prohibits network access and modifications to test files. This standardizes the task-facing message across harnesses. We do not attempt to equalize each harness's hidden system prompt, tool schema, parser hints, memory policy, or stopping rule; these remain part of the harness design and are therefore part of the experimental variable.

\textbf{Patch and evaluation contract.}
Candidate solutions are collected from repository state rather than from an agent's final message. After a harness terminates, times out, or returns an error, the runner extracts the diff against the base commit, removes known non-solution artifacts, and writes a SWE-bench-compatible prediction. This centralized patch contract makes heterogeneous harnesses comparable even when their native outputs differ: JSON, plain text, natural-language summaries, or no structured response all reduce to the same evaluator-facing patch format. Then, evaluation is performed with the official SWE-bench harness~\citep{jimenez_swebench_2024}.

%% file: sections/03_lite.tex
\section{Claw-SWE-Bench Lite}
\label{sec:lite}

The full 350-instance benchmark (full-350) is used for standard evaluation in this paper, but
rerunning all tasks at every development iteration is costly. A full run involves substantial token
consumption, API costs, wall-clock time, and log inspection effort. During adapter debugging, prompt
editing, model replacement, or regression testing, repeated full runs can make evaluation a major
development bottleneck. We therefore construct Claw-SWE-Bench Lite as a low-cost companion subset.
Its 80 instances are selected to approximate the overall and per-language Pass@1, relative harness
performance, and cost characteristics of full-350. Researchers can use Lite to quickly assess system
changes and return to full-350 at key stages for final reporting.

\subsection{Lite Subset Definition}
\label{sec:lite-def}

Lite-80 selects 10 instances from each of the eight language groups in full-350. It contains 70
non-Python instances from SWE-bench-Multilingual and 10 Python instances from
SWE-bench-Verified-Mini, matching the sources used in the full benchmark. In addition to balancing
languages, Lite enforces fixed difficulty-quartile quotas within each language group. It selects 2,
3, 3, and 2 instances from $Q_1$, $Q_2$, $Q_3$, and $Q_4$, respectively. This prevents the subset
from concentrating on unusually easy or difficult tasks. The final subset covers 38 of the 43
repositories in the full benchmark, or approximately 88\%, retaining broad repository coverage.
We use the full-350 evaluation results of seven harnesses paired with three models as calibration data for Lite subset selection.

\begin{figure}[t]
  \centering
  \begin{minipage}[t][\dimexpr 0.32\linewidth + 14pt\relax][t]{0.32\linewidth}
    \vspace*{0pt}\centering
    \includegraphics[width=\linewidth]{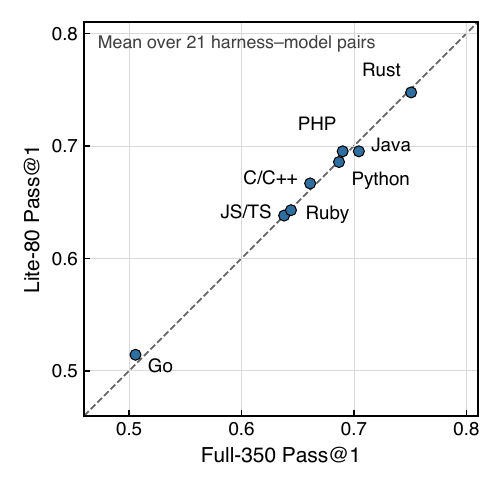}
    \vfill
    {\footnotesize (a) Per-language parity}
  \end{minipage}\hfill
  \begin{minipage}[t][\dimexpr 0.32\linewidth + 14pt\relax][t]{0.32\linewidth}
    \vspace*{0pt}\centering
    \includegraphics[width=\linewidth]{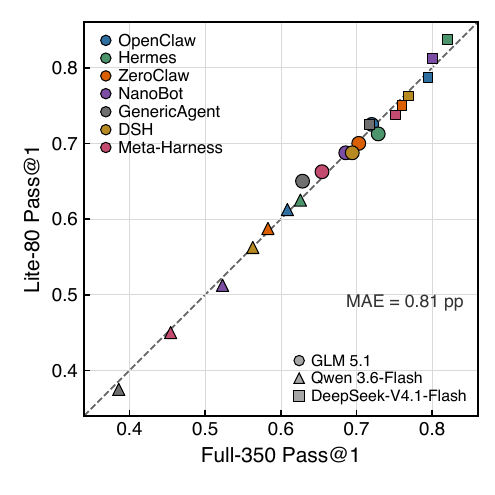}
    \vfill
    {\footnotesize (b) Cross-claw parity}
  \end{minipage}\hfill
  \begin{minipage}[t][\dimexpr 0.32\linewidth + 14pt\relax][t]{0.32\linewidth}
    \vspace*{0pt}\centering
    \includegraphics[width=\linewidth]{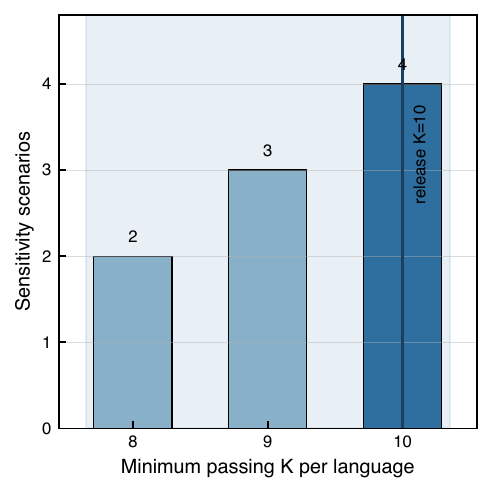}
    \vfill
    {\footnotesize (c) K-sweep sensitivity envelope}
  \end{minipage}
  \vspace{-1mm}
  \caption{Lite-80 parity with full-350. \textbf{(a)} Per-language Pass@1 averaged uniformly over 21 pairs. \textbf{(b)} Overall Pass@1 for 7 claws $\times$ 3 models. \textbf{(c)} K-sweep sensitivity envelope on the 21-column calibration pool; the minimum acceptable $K$ falls in $[8,10]$ across scenarios.}
  \vspace{-3mm}
  \label{fig:dist}
\end{figure}

\subsection{Subset Selection Method}
\label{sec:lite-method}

\paragraph*{Instance selection.}
TinyBenchmarks~\citep{tinybenchmarks} estimates full-benchmark performance using a small set of
representative examples. Following this idea, we formulate Lite construction as a constrained subset
selection problem. For Lite-80, we select 10 instances per language group under the 2/3/3/2
difficulty-quartile quotas. We compute the mean resolved rate of each instance across the 21
calibration columns and use these values to define quartiles within each language group. This
ensures that the selected instances cover different levels of relative difficulty.

The objective combines resolved-rate parity and pairwise ranking, with cost parity enforced during candidate screening. The
resolved-rate parity term minimizes the $L_1$ difference between Lite and full-350 over the
$21\times 8$ grid of calibration columns and language groups. However, small errors in individual
scores do not necessarily preserve the relative ordering of systems. We therefore include a pairwise
ranking hinge term~\citep{joachims_ranking_2002}. Within each language group, this term considers pairs of calibration columns
whose full-set resolved rates differ by more than 0.05. It applies a penalty if Lite reverses the
ordering or produces a gap below 0.025 in the direction of the full-set ordering. The objective thus
accounts for both absolute system performance and relative rankings.

Cost parity is enforced by filtering candidate subsets using the log-space difference between
Lite and full-350 mean per-instance costs. We require the mean absolute log error to be at most
0.10 on both evaluable columns and columns with complete cost records. This controls relative
cost deviations while reducing the number of evaluated instances.
We optimize the objective independently for each language group using local search with multiple
random restarts~\citep{marti_multistart_2013}. Each step swaps one selected instance with one unselected instance from the same
difficulty quartile. This preserves the instance count and difficulty quotas throughout the search.

\begin{table}[t]
  \centering
  \footnotesize
  \renewcommand{\arraystretch}{1.0}
  \setlength{\tabcolsep}{4pt}
  \caption{Comparison between the bare adapter and the full adapter. Both use the same GLM 5.1 backbone. All metrics are averaged over three runs per configuration. \textbf{Apply Failed} is the fraction of instances whose submitted patch cannot be applied to the repository by the SWE-bench evaluator.}
  \label{tab:adapter_diagnostic}
  \begin{tabular*}{\linewidth}{@{\extracolsep{\fill}}lccc@{}}
    \toprule
    \textbf{Configuration} & \textbf{Resolved} & \textbf{Pass@1(\%)} & \textbf{Apply Failed} \\
    \midrule
    Bare adapter & 67.7/350 & 19.3 & 69.1\%\\
    Full adapter & 249.3/350 & 71.2 & $<1.5\%$ \\
    \bottomrule
  \end{tabular*}
  \vspace{-5mm}
\end{table}

\paragraph*{Subset size selection.}
We determine the subset size through a $K$-sweep on the 21-column calibration pool. We vary $K$,
the number of instances per language group, under different ranking margins, restart counts,
random seeds, and mirror-parity settings. The minimum acceptable size ranges from 8 to 10
instances per language group: two scenarios pass at $K=8$, three require $K=9$, and four
require $K=10$. We select the largest of these scenario-specific minima, $K=10$,
giving 80 instances across the eight language groups.

\subsection{Validation Results}
\label{sec:lite-validation}

Across the 21 calibration columns, the mean Pass@1 is 0.6610 on full-350 and 0.6607 on Lite-80, a
difference of approximately 0.02 percentage points. Per-language deviations are all within
1 percentage point. We also compare Lite and full-350 across seven harnesses paired with
three shared models. The mean absolute Pass@1 difference across these 21 pairs is
0.81 percentage points. These results
show that Lite-80 closely matches full-benchmark success rates on harness--model pairs used for
calibration, covering both model and harness variation.

We also assess cost parity. Across the 11 harness--model pairs with complete per-instance cost estimates,
a Lite run costs on average 25.3\% of a full-350 run. The cost reduction
therefore comes mainly from evaluating fewer instances rather than selecting unusually cheap tasks.
Lite-80 can thus provide feedback for debugging and regression testing at roughly one quarter of the
evaluation cost.

%% file: sections/04_experimental_setup.tex
\section{Experimental Setup}
\label{sec:experiments}

We use Claw-SWE-Bench to compare a representative set of recent agent harnesses with different
designs. We focus on the capabilities and efficiency of general-purpose harnesses on software
engineering tasks. First, we compare the bare and full adapters to assess the role of the adapter
protocol in reliable task execution and patch submission. We then conduct the main comparison of
seven harnesses with three underlying models on full-350. We further validate Lite-80 on six
harness and model combinations excluded from subset construction and calibration. Using the main
experimental results, we examine the trade-off between task performance and running cost. Finally,
we assess the impact of future-commit leakage in some SWE-bench-Multilingual evaluation environments
by comparing results before and after cleanup.

\paragraph*{Claws.}
The main experiment includes seven harnesses: OpenClaw~\citep{steinberger_openclaw},
Hermes Agent~\citep{nous_hermes_agent}, nanobot~\citep{hkuds_nanobot},
ZeroClaw~\citep{zeroclaw_labs}, GenericAgent~\citep{generic_agent_2026},
DeepSeek Harness~\citep{deepseek_harness},
and Meta-Harness~\citep{lee_meta_harness_2026}, a method for automatically evolving harnesses.
For Meta-Harness, we use GenericAgent as the initial harness because its Python implementation
allows unrestricted code modification and is comparable in code size to the original seed harnesses.
We use DeepSeek-V4.1-Flash as the proposer model. We evaluate the evolved
harness separately from the original GenericAgent. We also include OpenSquilla~\citep{opensquilla_2026} in the Lite held-out
validation.

\begin{table}[t]
  \centering
  \footnotesize
  \renewcommand{\arraystretch}{0.95}
  \setlength{\tabcolsep}{4pt}
  \caption{Performance across seven harnesses with three models on Claw-SWE-Bench. Resolved is the number of solved instances. Cost, In, and Out are totals over 350 instances; In excludes cache-read tokens. Dur is mean per-instance duration, and Cache is the token-weighted mean cache hit rate. Bold indicates the highest Resolved and Pass@1 within each model group. $^{\dagger}$\,Based on GenericAgent.}
  \label{tab:b}
  \begin{tabular*}{\linewidth}{@{\extracolsep{\fill}}l|cc|ccccc@{}}
    \toprule
    \textbf{Harness} & \textbf{Resolved} & \textbf{Pass@1 (\%)} & \textbf{Cost (USD)} & \textbf{Dur (s)} & \textbf{In (M)} & \textbf{Out (M)} & \textbf{Cache (\%)} \\
    \midrule
    \multicolumn{8}{l}{\textit{Model: GLM 5.1}} \\
    OpenClaw & 249.3 & 71.2 & 338 & 511 & 68.7 & 8.1 & 92.0 \\
    Hermes Agent & \textbf{256.0} & \textbf{73.1} & 389 & 800 & 64.4 & 6.9 & 94.1 \\
    nanobot & 240.3 & 68.7 & 525 & 956 & 150.4 & 7.8 & 87.7 \\
    ZeroClaw & 244.0 & 69.7 & 391 & 548 & 93.3 & 4.3 & 90.9 \\
    GenericAgent & 220.3 & 62.9 & 86 & 576 & 33.1 & 5.1 & 66.8 \\
    DeepSeek Harness & 241.7 & 69.1 & 513 & 916 & 93.2 & 5.3 & 93.7 \\
    Meta-Harness$^{\dagger}$ & 230.0 & 65.7 & 190 & 379 & 79.7 & 5.7 & 72.2 \\
    \midrule
    \multicolumn{8}{l}{\textit{Model: Qwen 3.6-flash}} \\
    OpenClaw & 213.0 & 60.9 & 87 & 639 & 34.6 & 9.5 & 97.4 \\
    Hermes Agent & \textbf{219.7} & \textbf{62.8} & 104 & 639 & 44.5 & 7.3 & 97.4 \\
    nanobot & 183.7 & 52.5 & 175 & 636 & 466.1 & 9.9 & 65.1 \\
    ZeroClaw & 204.0 & 58.3 & 69 & 429 & 32.7 & 6.1 & 96.9 \\
    GenericAgent & 134.7 & 38.5 & 39 & 321 & 73.8 & 7.5 & 71.5 \\
    DeepSeek Harness & 198.7 & 56.8 & 91 & 614 & 38.4 & 9.7 & 97.2 \\
    Meta-Harness$^{\dagger}$ & 158.0 & 45.1 & 49 & 301 & 105.8 & 7.6 & 68.3 \\
    \midrule
    \multicolumn{8}{l}{\textit{Model: DeepSeek-V4.1-Flash}} \\
    OpenClaw & 279.3 & 79.8 & 50 & 566 & 38.5 & 25.4 & 97.3 \\
    Hermes Agent & \textbf{286.0} & \textbf{81.7} & 36 & 492 & 41.8 & 13.3 & 96.9 \\
    nanobot & 278.7 & 79.6 & 78 & 533 & 184.9 & 13.6 & 84.2 \\
    ZeroClaw & 266.3 & 76.1 & 99 & 1432 & 57.6 & 62.2 & 95.3 \\
    GenericAgent & 250.3 & 71.5 & 12 & 839 & 11.7 & 6.8 & 88.1 \\
    DeepSeek Harness & 267.7 & 76.5 & 41 & 1577 & 56.4 & 12.9 & 96.3 \\
    Meta-Harness$^{\dagger}$ & 262.3 & 74.9 & 13 & 1005 & 14.0 & 7.2 & 86.6 \\
    \bottomrule
  \end{tabular*}
  \vspace{-5mm}
\end{table}

\paragraph*{Models.}
The main experiment uses GLM 5.1~\citep{zai_glm_51},
Qwen 3.6-flash~\citep{alibaba_qwen36_flash}, and
DeepSeek-V4.1-Flash~\citep{deepseek_v41_flash}. Each model is paired with
all seven harnesses on full-350, giving 21 harness--model pairs. We hold the underlying model fixed within
each comparison to examine differences in harness performance.

\paragraph*{Evaluation metrics.}
Our primary metric is Pass@1, defined as the fraction of task instances marked as Resolved by the SWE-bench evaluator~\citep{jimenez_swebench_2024}. We also report total API cost, mean wall-clock duration, token usage,
and cache hit rate. API costs are calculated from the per-call token usage recorded by each harness and the fixed unit prices listed in Appendix~\ref{appendix:model-pricing}. Wall-clock
duration includes remote API latency. The adapter diagnostic additionally reports the patch
application failure rate. For Lite validation, we compare the full-350 and Lite-80 Pass@1 scores
of six held-out harness--model pairs. The diagonal represents equal scores on the two
task sets.

\paragraph*{Runtime configuration.}
Each instance runs in its corresponding SWE-bench Docker image. The repository is located at
\texttt{/testbed} and reset to the specified base commit. Standard evaluations apply future-commit
cleanup to the Multilingual instances. Unless explicitly varied in the adapter or cleanup
diagnostics, all runs share the task-prompt template, patch collector, evaluator, and result
aggregation code. We set a per-instance wall-clock timeout of 3,600 seconds and fix worker
concurrency at three. For both the adapter comparison and the main comparison of 21 harness--model
pairs, we evaluate each configuration on full-350 in three separate runs and report the mean
results. Other experiments use a single run per configuration on each task set. All model inference is performed through remote APIs. For the six
held-out harness--model pairs, we compare existing full-350 results with separate runs on the fixed
Lite-80 subset.

%% file: sections/05_results.tex
\section{Results}\label{sec:results}

\subsection{The Adapter Makes a General Agent Scorable}\label{sec:adapter_diagnostic}

We first test whether the adapter is merely an engineering wrapper or a necessary condition for OpenClaw to be reliably scored by SWE-bench. Table~\ref{tab:adapter_diagnostic} compares the same GLM 5.1 backbone under the bare adapter and the full adapter. The bare adapter places OpenClaw in the SWE-bench Docker environment and sends the task. The full adapter instead lets the model edit repository files through tools and has the runner export the patch from Git state.

The results show that minimal access is insufficient to create a reliable SWE-bench evaluation target. The bare adapter reaches only $19.3\%$ resolved rate. The main bottleneck is not that the model cannot edit code at all, but the fragility of directly generating unified-diff text: line numbers, context, hunk headers, or trailing newlines can make the patch fail to apply. The full adapter shifts the output responsibility from ``the model writes patch text'' to ``the model edits repository files and the runner exports the patch,'' reducing apply failures below $1.5\%$ and raising resolved rate to $71.2\%$. The following experiments therefore measure model and claw differences under a unified scoring contract, rather than testing whether a native agent can hand-write a SWE-bench-compatible diff.

\subsection{Performance across Harnesses}\label{sec:variation_harnesses}

To isolate the effect of harness choice, we hold the underlying model fixed and compare seven
harnesses under the shared Claw-SWE-Bench protocol. We conduct this comparison with GLM 5.1,
Qwen 3.6-flash, and DeepSeek-V4.1-Flash, yielding 21 harness--model pairs. This design allows us to examine
whether relative harness performance remains consistent across models or depends on the underlying
model. Each pair is evaluated on all 350 instances in three separate runs, giving 22,050 task executions in total. 
The main table reports resolved instance counts, Pass@1, total API cost, mean per-instance wall-clock duration, total uncached input tokens, total output tokens, and cache hit rate. 
All metrics are averaged over the three runs for each pair.

Table~\ref{tab:b} shows substantial differences in Pass@1 across harnesses even when the underlying
model is fixed. The largest gap reaches 24.3 percentage points(pp) on Qwen 3.6-flash. Specifically,
Hermes Agent achieves the highest mean Pass@1 with all three models, while some other harnesses
show more variation in their relative performance. For example, nanobot trails the best harness by
10.3~pp with Qwen 3.6-flash, but by only 2.1~pp with
DeepSeek-V4.1-Flash. 
Evolving GenericAgent through Meta-Harness also improves Pass@1 across all models. Following the corresponding experimental setup in the original Meta-Harness paper, we use the same instances for evolution and final evaluation. Meta-Harness uses evaluation feedback from these tasks during evolution, so comparisons with other harnesses are not conducted under equal conditions. Its performance gains should therefore not be interpreted as evidence of superiority in a fair comparison.
These results show that harness design affects how effectively model capabilities translate into successful issue resolution. They motivate further study of harness optimization and its interaction with the underlying model.

\Needspace{23\baselineskip}
\subsection{Efficiency and Cost Analysis}\label{sec:cost_perf}

\begin{wrapfigure}{r}{0.48\textwidth}
  \vspace{-\intextsep}
  \centering
  \addtolength{\abovecaptionskip}{-5mm}
  \includegraphics[width=\linewidth]{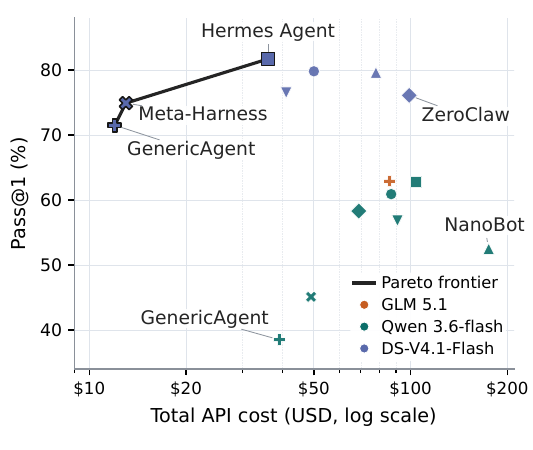}
  \caption{Resolve-rate--cost Pareto frontier for part of the results. The black line connects non-dominated points.}
  \label{fig:pareto_frontier}
  \vspace{-5mm}
\end{wrapfigure}

We further analyze the trade-off between task success and API cost using the main results in
Table~\ref{tab:b}. Figure~\ref{fig:pareto_frontier} plots part of the results with total API cost on the
horizontal axis and Pass@1 on the vertical axis. Each point represents one harness and model
combination evaluated on full-350, with both metrics averaged over three runs. We highlight the
Pareto frontier across all 21 harness--model pairs.

As shown in Figure~\ref{fig:pareto_frontier}, DeepSeek-V4.1-Flash, the newest of the three evaluated
models, generally achieves higher Pass@1 than the other two models. Although it typically generates
more output tokens, its low cached-input token price (Appendix~\ref{appendix:model-pricing}) helps keep total API costs lower for most
harnesses. The Pareto frontier consists of three harness--model pairs, all using DeepSeek-V4.1-Flash:
GenericAgent, Meta-Harness, and Hermes Agent. GenericAgent provides the lowest-cost point, while
Hermes Agent reaches the highest Pass@1 of 81.7\% at \$36. These pairs offer different
performance and cost trade-offs under the same model, showing that harness choice affects not only
task success but also the cost of agent execution.

\Needspace{20\baselineskip}
\subsection{Evaluation of Claw-SWE-Bench Lite}\label{sec:lite_heldout}

\begin{wrapfigure}{r}{0.32\textwidth}
  \vspace{-\intextsep}
  \centering
  \addtolength{\abovecaptionskip}{-4mm}
  \includegraphics[width=\linewidth]{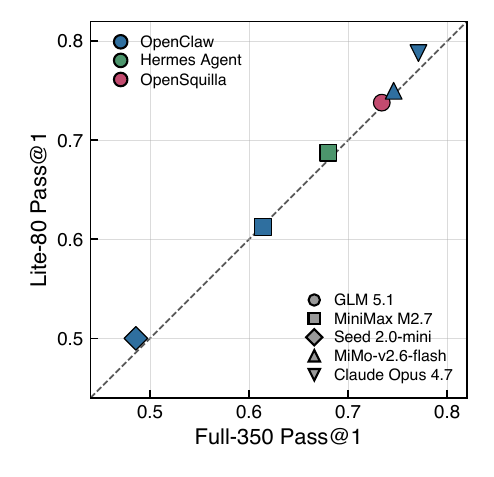}
  \caption{Pass@1 on full-350 and Lite-80 for six harness--model pairs. The dashed line indicates equal scores.}
  \label{fig:lite_heldout_parity}
  \vspace{-2mm}
\end{wrapfigure}

We further assess whether Lite-80 can approximate full-benchmark
performance on systems excluded from calibration. We evaluate the fixed subset on six new harness--model pairs: OpenSquilla with GLM 5.1, OpenClaw and Hermes Agent with
MiniMax M2.7~\citep{minimax_m27}, and OpenClaw with
Seed 2.0-mini~\citep{bytedance_seed_20_mini}, MiMo-v2.6-flash and Opus 4.7~\citep{anthropic_claude_opus_47}. These pairs include four new models that are absent from the calibration pool, while OpenSquilla introduces a new harness. None of these
pairs is used for subset construction or calibration. 

Following Figure~\ref{fig:dist}(b), Figure~\ref{fig:lite_heldout_parity} compares full-350 Pass@1 on
the horizontal axis with Lite-80 Pass@1 on the vertical axis, with one point per pair.
The diagonal represents equal scores on the two task sets; points closer to this line indicate
closer agreement. Across these six pairs, the maximum absolute difference is
1.70~pp, and their ranking is unchanged between full-350 and Lite-80.

\subsection{Effect of Future-Commit Cleanup}\label{sec:leak_fix}

During manual inspection of the task environments, we found that SWE-bench-Multilingual
containers retained Git history after the specified base commit, including the reference fixes.
The image-building code used in these runs cleaned future history for Python repositories but
not for non-Python repositories. For the latter, removing the remote origin left the local
history intact. Agents could therefore use pull request identifiers from task descriptions to
locate repair commits and retrieve reference patches.

\Needspace{24\baselineskip}
\begin{wrapfigure}{r}{0.4919\textwidth}
  \vspace{-\intextsep}
  \vspace{-1mm}
  \centering
  \addtolength{\abovecaptionskip}{-1mm}
  \includegraphics[width=\linewidth]{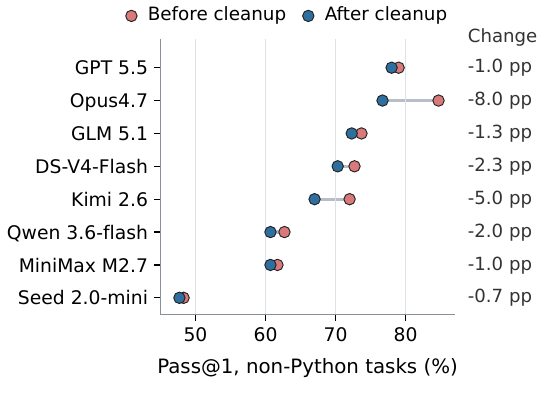}
  \caption{Effect of future-commit cleanup.}
  \label{fig:leak_fix_openclaw_multilingual}
\end{wrapfigure}

We address this issue during workspace preparation by removing accessible Git history beyond the base commit before agent execution. To measure the impact of cleanup, we fix OpenClaw and compare eight models before and after this change on the 300 Multilingual instances. As shown in Figure~\ref{fig:leak_fix_openclaw_multilingual}, Pass@1 after cleanup is no higher than before cleanup for any of the eight models. The magnitude of the change varies across models. Opus 4.7 shows the largest decrease, at 8.0~pp. Kimi 2.6 drops by 5.0~pp, and Qwen 3.6-flash by 2.0~pp.
We therefore apply future-commit cleanup in standard evaluations to prevent direct access to reference fixes and support fair comparisons.

%% file: sections/07_conclusion.tex
\section{Conclusion}\label{sec:discussion}

We introduced Claw-SWE-Bench and its 80-instance companion subset, Claw-SWE-Bench Lite. The shared adapter protocol connects general-purpose agents to the SWE-bench execution environment, patch submission process, and scoring system. This enables consistent, reproducible evaluation of their software engineering capabilities. We evaluated seven harnesses with three models using the same task prompts, outer runtime budgets, and task set. The results show that general-purpose harnesses can effectively complete real-world software engineering tasks. Even with the underlying model fixed, task success rates and running costs vary substantially across harnesses. SWE-style coding agent evaluation should therefore treat the harness as a key controlled variable rather than an implementation detail hidden behind model scores. We hope the full benchmark and Lite subset will provide stable reference points for future research. They support comparisons of new and evolved harnesses under the same tasks, budgets, and cost-accounting conventions. They also make low-cost debugging and replication easier to incorporate into routine research workflows.

%% file: appendix_content.tex
\appendix
\numberwithin{figure}{section}
\numberwithin{table}{section}
\renewcommand{\theHfigure}{appendix.\thesection.\arabic{figure}}
\renewcommand{\theHtable}{appendix.\thesection.\arabic{table}}
\raggedbottom
\input{appendix/C_reproducibility}
\input{appendix/D_compute_env}
\input{appendix/meta_harness_implementation}
\input{appendix/F_lite_construction}

\input{appendix/G_per_repo_breakdown}
\input{appendix/I_license_ethics}
\input{appendix/J_llm_comparison}

%% file: appendix/C_reproducibility.tex
\section{Reproducibility Statement}\label{appendix:reproducibility}

This appendix consolidates the artefacts and protocol required to reproduce every cell of \S\ref{sec:variation_harnesses}.

\subsection{Code release}

\paragraph{Harness adapters.} Adapters for seven claws---OpenClaw, Hermes Agent, ZeroClaw, nanobot, GenericAgent, DeepSeek Harness, and Meta-Harness---are released. The release includes \verb|run_infer.py| and \verb|run_eval.py| together with the per-harness orchestrator, agent adapter, and workspace modules.

\subsection{Data release}

\textbf{Claw-SWE-Bench} (350 instance IDs plus metadata) and \textbf{Claw-SWE-Bench-Lite} (80 instance IDs with cost-aware selection metadata) are released as JSON files. The underlying issues and repositories are hosted on the upstream SWE-bench-Multilingual and SWE-bench-Verified-Mini sources.

\subsection{Reproduction protocol}

\paragraph{Per-instance run.}
\begin{verbatim}
python3 run_infer.py \
    --claw <claw_name> \
    --dataset multilingual \
    --instance_file config/multilingual_300_instances.txt \
    --model <provider/model_name> \
    --run_id <run_label> \
    --timeout 3600 \
    --workers 3
\end{verbatim}

\paragraph{Evaluation.}
\begin{verbatim}
python3 run_eval.py \
    --predictions artifacts/<run_id>/predictions.jsonl \
    --dataset_name SWE-bench/SWE-bench_Multilingual \
    --run_id <run_id> \
    --max_workers 8
\end{verbatim}

The commands above cover the 300 Multilingual instances. For the 50 Verified-Mini instances,
use \texttt{--dataset verified} and \texttt{--instance\_file}
\texttt{config/verified\_mini\_50.txt} for inference, and \texttt{--dataset\_name}
\texttt{princeton-nlp/SWE-bench\_Verified} for evaluation, with a separate run ID.

All seven claws share these inference and evaluation entry points. Claw-specific model settings
are documented in the released README; Meta-Harness uses
\texttt{--claw generic --candidate <candidate\_dir>}.

\subsection{Model Pricing and API Cost Accounting}
\label{appendix:model-pricing}

We calculate API costs using fixed unit prices and the per-call token usage recorded by each harness.
For a given model, the same prices apply across all harnesses. We account separately for uncached
input, cached input, and output tokens. Table~\ref{tab:model-pricing} lists the prices used for the
three models in the main experiment. All prices are in USD per million tokens.

\begin{table}[htbp]
\centering
\small
\setlength{\tabcolsep}{4pt}
\caption{API prices for the three models used in the main experiment (USD per million tokens).}
\label{tab:model-pricing}
\begin{tabularx}{\linewidth}{@{}lrrr>{\raggedright\arraybackslash}X@{}}
\toprule
\textbf{Model} & \shortstack{\textbf{Uncached}\\\textbf{input}} & \shortstack{\textbf{Cached}\\\textbf{input}} & \textbf{Output} & \textbf{Source} \\
\midrule
GLM-5.1 & 1.40 & 0.26 & 4.40 & Z.ai official pricing page \\
Qwen 3.6-flash & 0.25 & 0.05 & 1.50 & Alibaba Cloud Model Studio international pricing page (Singapore) and Context Cache documentation \\
DeepSeek-V4.1-Flash & 0.30 & 0.006 & 1.20 & DeepSeek official peak-hour pricing \\
\bottomrule
\end{tabularx}
\end{table}

\Needspace{6\baselineskip}
\paragraph{Qwen pricing tier and cache pricing.}
We use the Qwen 3.6-flash pricing tier for requests with at most 256K input tokens. The official
model pricing table does not list a separate price for cached input. The Context Cache documentation
specifies that implicit cache hits are charged at 20\% of the standard input price. We therefore use
a cached-input price of $0.25 \times 20\% = \text{US\$0.05}$ per million tokens.

\paragraph{DeepSeek peak and off-peak pricing.}
Peak hours for DeepSeek-V4.1-Flash are 01:00 to 04:00 and 06:00 to 10:00 UTC on weekdays. Prices
are halved at all other times. The off-peak prices for uncached input, cached input, and output are
US\$0.15, US\$0.003, and US\$0.60 per million tokens, respectively. We use the peak-hour prices
in Table~\ref{tab:model-pricing} for all cost calculations, without off-peak discounts. This makes
cost accounting independent of when API calls are made.

\paragraph{Cost calculation.}
The cost of each API call is calculated as
\[
C = \frac{
  N_{\mathrm{fresh}}p_{\mathrm{in}}
  + N_{\mathrm{cache}}p_{\mathrm{cache}}
  + N_{\mathrm{out}}p_{\mathrm{out}}
}{10^{6}}.
\]
Here, $N_{\mathrm{fresh}}$, $N_{\mathrm{cache}}$, and $N_{\mathrm{out}}$ denote the numbers of uncached
input, cached input, and output tokens, respectively. The corresponding unit prices are
$p_{\mathrm{in}}$, $p_{\mathrm{cache}}$, and $p_{\mathrm{out}}$, as listed in Table~\ref{tab:model-pricing}.
Cached input tokens are excluded from $N_{\mathrm{fresh}}$ to avoid double counting. The cost of a
full evaluation run is the sum of all API call costs across all task instances. Costs in the main
table are first summed within each full run and then averaged over the three runs.

%% file: appendix/D_compute_env.tex
\section{ Compute and Environment Details}\label{appendix:compute}

This appendix documents the hardware, run-time parameters, and aggregate
wall-clock cost of the experiments reported in Section~\ref{sec:experiments}. All seven
claws (\textsc{openclaw}, \textsc{hermes-agent}, \textsc{zeroclaw}, \textsc{nanobot},
\textsc{generic}, \textsc{deepseek harness}, and \textsc{meta-harness})
were executed on a single host with identical run-time parameters; the claw
implementation is the only experimental variable.

\Needspace{14\baselineskip}
\subsection{Hardware}

\begin{table}[htbp]
\centering
\small
\begin{tabular}{ll}
\toprule
\textbf{Item} & \textbf{Value} \\
\midrule
Architecture & x86\_64 \\
OS           & Linux 6.8.0-106-generic (Ubuntu) \\
CPU cores    & 16 \\
RAM          & 61~GiB total \\
GPU          & None (model inference is API-only) \\
\bottomrule
\end{tabular}
\caption{Host hardware. Model inference runs on remote provider APIs, so no local
GPU is required; the host is used only for harness orchestration, Docker
containers, and patch evaluation.}
\label{tab:appx-d-hardware}
\end{table}

\Needspace{14\baselineskip}
\subsection{Run-time parameters (held equal across all seven claws)}

\begin{table}[htbp]
\centering
\small
\begin{tabular}{ll}
\toprule
\textbf{Parameter} & \textbf{Value} \\
\midrule
Per-instance timeout                                & 3600~s (60~min) \\
Repeats per instance                                & 3 \\
Concurrency                                         & 3 threads \\
Shell tool timeout (each harness)                   & 180~s \\
\bottomrule
\end{tabular}
\caption{Run-time parameters for the main harness comparison. These are overridden via CLI flags so that all seven
claws see the same per-instance budget; the prompt template is also identical across claws.}
\label{tab:appx-d-runtime}
\end{table}

\subsection{API provider}

All model inference is routed through OpenRouter. \textbf{API keys are NOT
included in any released artifact}.

\subsection{ Compute cost (aggregate)}

The 21 full-350 runs used for Lite calibration cover seven claws paired with
three models, totaling 7{,}350 instance executions. Their summed per-instance
durations amount to approximately \textbf{1{,}430 hours}. This total excludes
the additional OpenClaw model sweep, the adapter diagnostic
(\S\ref{sec:adapter_diagnostic}), the pre-cleanup runs
(\S\ref{sec:leak_fix}), and Lite-80 validation runs.

%% file: appendix/meta_harness_implementation.tex
\section{Meta-Harness Implementation Details}\label{appendix:meta_harness_implementation}

We adapt the original outer-loop implementation of Meta-Harness
to Claw-SWE-Bench and perform a single evolution round starting from GenericAgent.
The resulting harness is reported as Meta-Harness in Table~\ref{tab:b}.
The benchmark adapter and evaluation protocol remain fixed throughout this process.

\paragraph*{Seed harness and search space.}
We initialize evolution from GenericAgent, an open-source Python harness with approximately
3,700 lines of code. The candidate contains a complete copy of the harness source tree.
The proposer can modify the agent loop, tools, system prompts, and context management.
The evaluation adapter, model configuration, and instance list are excluded from modification.

\paragraph*{Proposer and feedback.}
We use DeepSeek-V4.1-Flash as the proposer model, with reasoning effort set to \texttt{xhigh}.
The proposer operates in an isolated working directory and can inspect harness source code,
per-instance scores, and full execution traces. These artifacts include \texttt{agent\_stdout.log},
\texttt{git.patch}, and evaluation outputs such as \texttt{report.json}.
It produces one candidate and a falsifiable hypothesis about the proposed change.
We adapt the original SKILL instructions to the SWE-bench setting. The instructions require
the proposer to identify failure patterns in the traces before editing and to target a single
mechanism. They prohibit hard-coded instance identifiers, repository names, and other
task-specific content, as well as Git-history inspection and network retrieval.

\paragraph*{Search data and auditing.}
Following the TerminalBench-2 setting of Meta-Harness, we use all 350 Claw-SWE-Bench instances
for both evolution and final evaluation, without a held-out task split. To detect task-specific
modifications, we apply regular-expression checks to added code for instance identifiers and
repository names. We also manually inspect the complete candidate diff. These checks target
explicit task-specific code; they do not establish generalization to unseen tasks.

\paragraph*{Final evaluation and cross-model transfer.}
During evolution, both the proposer and the task-execution model use DeepSeek-V4.1-Flash.
After evolution, we freeze the resulting harness. We evaluate it with DeepSeek-V4.1-Flash
and transfer it to GLM 5.1 and Qwen 3.6-flash by changing only the execution model, without
further evolution. The proposer does not participate in these final task executions.
This experiment assesses cross-model transfer on the same benchmark tasks. The main-table
results follow the standard evaluation configuration, with three separate runs per model.

\paragraph*{Evolution cost.}
Proposal generation takes approximately eight minutes and costs less than US\$0.10 in API fees.
The full-350 evaluation of the candidate costs approximately US\$13 and takes about 13 hours
with eight concurrent workers.

%% file: appendix/F_lite_construction.tex
\section{Lite-80 Construction Detail}\label{appendix:lite}

This appendix records the released Lite-80 construction used in
Section~\ref{sec:lite}. The current release is built using full-350 results from 21
harness--model pairs spanning seven claws and three models.

\subsection{Calibration Pool and Constraints}

Lite selection is calibrated on 21 evaluation columns: seven claws
(OpenClaw, Hermes, ZeroClaw, NanoBot, GenericAgent, DSH, and Meta-Harness)
paired with GLM~5.1, Qwen~3.6-flash, and DeepSeek-V4.1-Flash. Within-model
claw comparisons use this $7\times 3$ grid.

Based on the $K$-sweep analysis described in Section~\ref{sec:lite-method},
we choose 10 instances per language, giving an 80-instance Lite subset.
Within a language, the 10 selected instances must follow the fixed
$2/3/3/2$ allocation over difficulty quartiles $Q_1/Q_2/Q_3/Q_4$.
Quartiles are computed from the mean resolved rate across the
calibration pool, so the strata reflect multi-model and multi-claw
difficulty rather than a single harness's behavior.

\subsection{Objective and Solver}

Let $x_i\in\{0,1\}$ indicate whether instance $i$ is selected.  The
selection loss combines two terms. First, a resolve-rate L1 term
matches Lite-implied and full-350 rates over the $21\times 8$ grid of
calibration column by language.  Second, a pairwise ranking hinge
penalizes column inversions: for column pairs whose full-set rates
differ by more than $\texttt{RANK\_EPS}=0.05$, the loss is active when
the Lite-predicted ordering is wrong or within a margin of $0.025$.
\subsection{K-sweep Decision}

Table~\ref{tab:lite-k-sweep} summarizes the $K$-sweep on the 21-column
calibration pool. The scan evaluates resolve-rate, cost, and operational
criteria under different sensitivity scenarios. The minimum acceptable sizes
range from 8 to 10 instances per language ($K^*\in[8,10]$). We select the
largest of these minima, $K=10$, yielding $8\times10=80$ instances.

\begin{table}[htbp]
\centering
\small
\caption{Subset-size sensitivity analysis on the 21-column calibration pool.
Each scenario row reports the smallest $K$ meeting the evaluation criteria.}
\label{tab:lite-k-sweep}
\begin{tabular}{llr}
\toprule
Scenario & Driver class & $K^*_{\mathrm{scenario}}$ \\
\midrule
main & -- & 9 \\
MARGIN\_0 & margin & 10 \\
MARGIN\_0.5overK & margin & 10 \\
restart\_50 & restart & 10 \\
restart\_500 & restart & 8 \\
seed\_offset\_1000 & seed & 8 \\
seed\_offset\_2000 & seed & 10 \\
mirror\_B & mirror parity & 9 \\
mirror\_A & mirror parity & 9 \\
\midrule
\textbf{release} & $K^*_{\max}$ & \textbf{10} \\
\bottomrule
\end{tabular}
\end{table}





\subsection{Coverage and Cost}

The released Lite-80 contains 38 unique repositories, covering approximately
$38/43=88\%$ of the repositories in the full benchmark. The mean Lite/full
run-cost ratio is $25.3\%$. This supports the intended use of Lite as an
approximately four-times cheaper evaluation surface for debugging,
regression testing, and preliminary model or claw comparisons.

%% file: appendix/I_license_ethics.tex
\section{ License and Ethics}\label{appendix:license}

\paragraph{License of the released artifact.}
This benchmark is derived from two upstream sources, both released under MIT.
\textbf{(i) SWE-bench Multilingual} is hosted at
\url{huggingface.co/datasets/SWE-bench/SWE-bench_Multilingual}, released as
part of the official SWE-bench project.
\textbf{(ii) SWE-bench Verified-Mini}
is an MIT-licensed subset of SWE-bench Verified (the OpenAI-validated
500-instance Python subset), hosted at
\url{github.com/mariushobbhahn/SWEBench-verified-mini}.

We retain both upstream LICENSE files and citations. The underlying source
code in each task instance retains the license of its original GitHub
repository. Users
redistributing patches or derivative work must comply with the
per-repository license. \texttt{REPO\_LICENSES.md} in the released artifact
lists every underlying repository and its current upstream license.

%% file: appendix/J_llm_comparison.tex
\section{Variation Along the LLM Axis}\label{sec:variation_llms}\label{appendix:llm_comparison}

\subsection{Aggregate Results}

We examine variation along the LLM axis using OpenClaw as the reference claw with eleven models on the full 350-instance set. The models are GPT 5.5, Claude Opus 4.7, GLM 5.1, DeepSeek-V4 Pro and Flash, Kimi K2.6, MiniMax M2.7, Qwen 3.6-flash, Seed 2.0-mini, MiMo-v2.6-flash, and LongCat-2.0. Table~\ref{tab:a2} reports aggregate results. The highest resolved rate is achieved by GPT 5.5, at $78.0\%$ (273/350), followed by Claude Opus 4.7 at $77.1\%$ (270/350). The lowest cell is Seed 2.0-mini, at $48.6\%$ (170/350). The observed Pass@1 spread across these OpenClaw runs is $29.4$\,pp.

Accuracy ranking, however, is not cost ranking. GPT 5.5 has the highest Pass@1, but its full 350-instance run costs $\$1399.1$; Claude Opus 4.7 is only $0.9$\,pp lower, with cost $\$1082.0$. By contrast, DeepSeek-V4 Pro reaches $71.7\%$ Pass@1 at total cost $\$81.3$, while DeepSeek-V4 Flash reaches $70.3\%$ at only $\$8.2$. Qwen 3.6-flash reaches $60.9\%$ Pass@1 at $\$77.5$; GLM 5.1 reaches $72.0\%$ at $\$290.6$.

\begin{table}[htbp]
  \centering
  \footnotesize
  \renewcommand{\arraystretch}{1}
  \setlength{\tabcolsep}{2pt}
  \caption{LLM-axis variation: OpenClaw $\times$ 11 models on full-350, with one run per model. Cost is total API cost (USD); Dur is mean duration per instance (seconds); In/Out are total non-cached input/output tokens (millions); Turns is mean turns; Cache is the token-weighted input cache hit rate (\%). Rows are grouped by model type and sorted by Pass@1 within each group. The best Pass@1 and lowest Cost are in \textbf{bold}.}
  \label{tab:a2}
  \begin{tabular*}{\linewidth}{@{\extracolsep{\fill}}ll|cccccccc@{}}
    \toprule
    \textbf{Model} & \textbf{Type} & \textbf{Resolved} & \textbf{Pass@1} & \textbf{Cost} & \textbf{Dur} & \textbf{In(M)} & \textbf{Out(M)} & \textbf{Turns} & \textbf{Cache} \\
    \midrule
    GPT 5.5             & Flagship & \textbf{273} & \textbf{78.0} & 1399.1 & 603.7 & 40.3 & 15.7 & 67.0 & 97.3 \\
    Claude Opus 4.7     & Flagship & 270 & 77.1 & 1082.0 & 424.6 & 35.6 & 6.2  & 61.6 & 97.0 \\
    GLM 5.1             & Flagship & 252 & 72.0 & 290.6  & 511.2 & 60.0 & 7.0  & 57.2 & 91.9 \\
    DeepSeek-V4 Pro     & Flagship & 251 & 71.7 & 81.3   & 662.3 & 19.3 & 11.0 & 47.1 & 97.4 \\
    LongCat-2.0         & Flagship & 235 & 67.1 & 478.0  & 1091.4 & 566.8 & 21.7 & 65.3 & 58.2 \\
    Kimi K2.6            & Flagship & 234 & 66.9 & 633.7  & 1235.3& 75.6 & 12.1 & 78.7 & 92.1 \\
    MiniMax M2.7        & Flagship & 215 & 61.4 & 196.7  & 1165.6& 25.0 & 9.0  & 94.8 & 96.2 \\
    \midrule
    MiMo-v2.6-flash     & Flash    & 261 & 74.6 & 22.3   & 1791.0 & 60.8 & 36.4 & 57.7 & 95.5 \\
    DeepSeek-V4 Flash   & Flash    & 246 & 70.3 & \textbf{8.2}   & 430.0 & 13.8 & 12.9 & 51.2 & 98.5 \\
    Qwen 3.6-flash      & Flash    & 213 & 60.9 & 77.5   & 638.9 & 31.7 & 8.8  & 68.2 & 97.3 \\
    Seed 2.0-mini       & Flash    & 170 & 48.6 & 19.4   & 1153.0& 89.8 & 21.6 & 44.4 & 79.4 \\
    \bottomrule
  \end{tabular*}
\end{table}

\subsection{Per-Language Resolved Rate}
\label{appendix:per-repo:g1}

Table~\ref{tab:appG-per-lang-A} expands the aggregate results in Table~\ref{tab:a2}
by language for the same eleven OpenClaw--model pairs. Models are listed
in descending order of overall resolved rate; each Total matches the corresponding
Pass@1 in Table~\ref{tab:a2}.

\begin{table}[htbp]
\centering
\small
\caption{Per-language resolved rate (\%) under OpenClaw for the 11 models in
Table~\ref{tab:a2}, using the same runs and accounting. Best language per model
in \textbf{bold}, worst \underline{underlined}.}
\label{tab:appG-per-lang-A}
\begin{tabular}{lcccccccc|c}
\toprule
Model & Java & Go & Rust & JS/TS & C/C++ & Ruby & PHP & Python & Total \\
\midrule
GPT 5.5           & 86.0          & \underline{61.9} & \textbf{93.0} & 79.1          & 81.0          & 70.5 & 74.4 & 78.0 & 78.0 \\
Claude Opus 4.7   & 86.0          & \underline{61.9} & \textbf{88.4} & 81.4          & 71.4          & 68.2 & 76.7 & 82.0 & 77.1 \\
MiMo-v2.6-flash   & \textbf{86.0} & \underline{64.3} & 81.4 & 72.1 & 78.6 & 70.5 & 74.4 & 70.0 & 74.6 \\
GLM 5.1           & 76.7 & \underline{59.5} & \textbf{81.4} & 69.8 & 76.2 & 68.2 & 74.4 & 70.0 & 72.0 \\
DeepSeek-V4 Pro   & \textbf{79.1} & \underline{50.0} & \textbf{79.1} & 74.4          & 73.8          & 72.7 & 76.7 & 68.0 & 71.7 \\
DeepSeek-V4 Flash & \textbf{81.4} & \underline{50.0} & 79.1          & 74.4          & 73.8          & 63.6 & 69.8 & 70.0 & 70.3 \\
LongCat-2.0       & 74.4 & \underline{54.8} & \textbf{81.4} & 67.4 & 61.9 & 63.6 & 67.4 & 66.0 & 67.1 \\
Kimi K2.6         & 69.8          & \underline{50.0} & \textbf{74.4} & 65.1          & 71.4          & 68.2 & 69.8 & 66.0 & 66.9 \\
MiniMax M2.7      & 65.1          & \underline{47.6} & \textbf{76.7} & 55.8          & 61.9          & 61.4 & 55.8 & 66.0 & 61.4 \\
Qwen 3.6-flash    & 60.5 & \underline{47.6} & \textbf{69.8} & 48.8 & 66.7 & 68.2 & 62.8 & 62.0 & 60.9 \\
Seed 2.0-mini     & \textbf{60.5} & \underline{33.3} & 55.8          & 44.2          & 54.8          & 36.4 & 48.8 & 54.0 & 48.6 \\
\bottomrule
\end{tabular}
\end{table}

\paragraph{Commentary.}
Java and Rust dominate the per-model maxima: Rust is the best language for 7 of
11 models, Java for 3, and the two tie for DeepSeek-V4 Pro. Go is the worst
language for all 11 models without exception (range 33.3--64.3\%), sitting
10--22~pp below each model's overall mean.